\documentclass[10pt,twocolumn,letterpaper]{article}

\usepackage[pagenumbers]{cvpr} 

\definecolor{cvprblue}{rgb}{0.21,0.49,0.74}
\usepackage[pagebackref,breaklinks,colorlinks,allcolors=cvprblue]{hyperref}
\usepackage{array,booktabs,makecell}
\newcolumntype{C}[1]{>{\centering\arraybackslash}m{#1}} 

\usepackage{algorithmic}
\usepackage{algorithm}

\usepackage{multirow}
\usepackage{float}
\usepackage{graphicx}
\usepackage{caption}
\usepackage{subcaption}
\usepackage{pifont}
\usepackage{graphicx}
\usepackage{cuted}   
\usepackage{capt-of}
\usepackage[table,xcdraw]{xcolor}
\usepackage{multirow}
\usepackage{booktabs}

\usepackage{colortbl}
\def\paperID{16974} 
\def\confName{CVPR}
\def\confYear{2026}

\title{Autoregressive Image Generation Needs Only a Few Lines of Cached Tokens}

\author{
    \textbf{Ziran Qin}$^1$\thanks{Equal contribution.} \quad \textbf{Youru Lv}$^1$$^*$ \quad \textbf{Mingbao Lin$^2$} \\ 
    \textbf{Zeren Zhang}$^3$ \quad \textbf{Chaofan Gan}$^1$\quad \textbf{Tieyuan Chen}$^1$ \quad \textbf{Weiyao Lin}$^1$\thanks{Corresponding author.}\\
    $^1$Shanghai Jiao Tong University \quad $^2$Rakuten \quad $^3$Peking University\\
    {\tt {\color{blue}Project: https://github.com/Zr2223/LineAR}}
}

\begin{document}
\maketitle

\begin{strip}
  \centering
    \vspace{-4em} 
  \includegraphics[width=\textwidth]{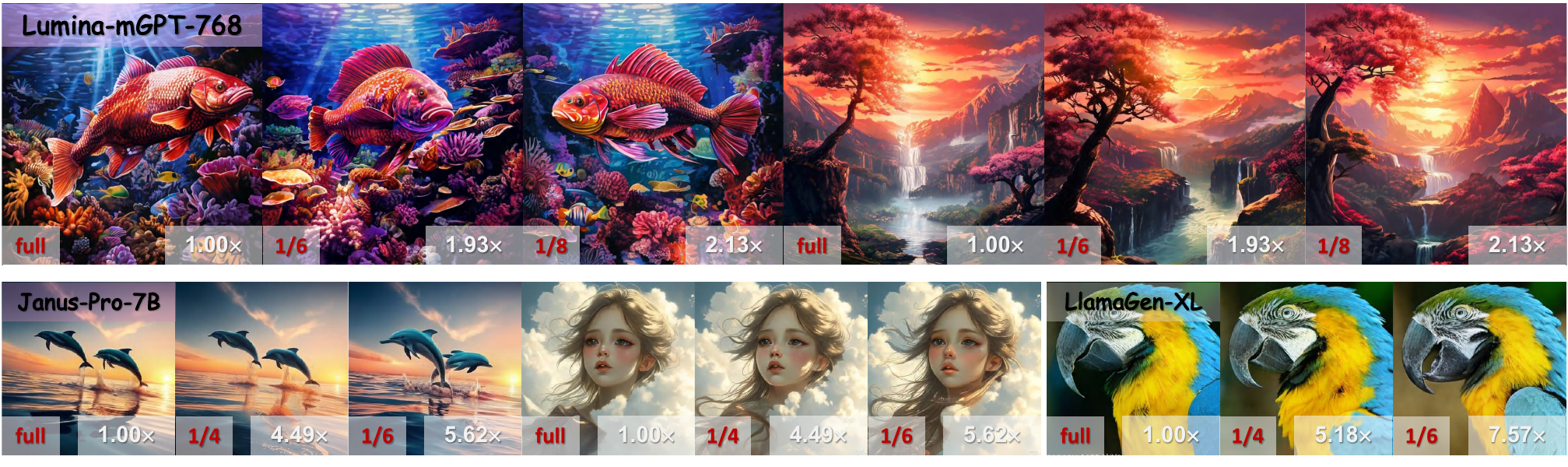}
  \vspace{-1.5em} 
  \captionof{figure}{%
   {LineAR enables efficient autoregressive image generation, preserving only 1/8, 1/6, and 1/6 of the KV cache, achieving up to 2.13$\times$, 5.62$\times$, and 7.57$\times$ speedup on Lumina-mGPT, Janus-Pro, and LlamaGen models, with improved or comparable generation quality.}%
  }
  \label{fig:teaser}
  \vspace{-0.2em}
\end{strip}

\begin{abstract}

Autoregressive (AR) visual generation has emerged as a powerful paradigm for image and multimodal synthesis, owing to its scalability and generality. However, existing AR image generation suffers from severe memory bottlenecks due to the need to cache all previously generated visual tokens during decoding, leading to both high storage requirements and low throughput. In this paper, we introduce \textbf{LineAR}, a novel, training-free progressive key-value (KV) cache compression pipeline for autoregressive image generation. 
By fully exploiting the intrinsic characteristics of visual attention, LineAR manages the cache at the line level using a 2D view, preserving the visual dependency regions while progressively evicting less-informative tokens that are harmless for subsequent line generation, guided by inter-line attention. 
LineAR enables efficient autoregressive (AR) image generation by utilizing only a few lines of cache, achieving both memory savings and throughput speedup, while maintaining or even improving generation quality. Extensive experiments across six autoregressive image generation models, including class-conditional and text-to-image generation, validate its effectiveness and generality. LineAR improves ImageNet FID from 2.77 to 2.68 and COCO FID from 23.85 to 22.86 on LlamaGen-XL and Janus-Pro-1B, while retaining only 1/6 KV cache. It also improves DPG on Lumina-mGPT-768 with just 1/8 KV cache.
Additionally, LineAR achieves significant memory and throughput gains, including up to 67.61\% memory reduction and 7.57$\times$ speedup on LlamaGen-XL, and 39.66\% memory reduction and 5.62$\times$ speedup on Janus-Pro-7B.

\end{abstract}

\section{Introduction}
\label{sec:intro}


Autoregressive (AR) models with the ``next-token'' prediction paradigm have achieved remarkable success in large language models (LLMs)~\cite{achiam2023gpt,liu2024deepseek} and have recently been extended to visual generation, producing high-quality results comparable to state-of-the-art diffusion models~\cite{sun2024autoregressive,he2024mars}. Their inherent scalability and generality also make AR models a promising foundation for multi-modal understanding and generation~\cite{wang2024emu3,liu2024lumina-mgpt}. Despite these advances, AR image generation suffers from inefficiencies during inference due to the greedy accumulation of a heavy KV cache from all generated tokens, leading to memory overhead that scales linearly with sequence length. For instance, generating a $1024 \times 1024$ image with Lumina-mGPT~\cite{liu2024lumina-mgpt} requires caching over 4K tokens, which not only leads to excessive memory consumption but also significantly slows down throughput, complicating practical deployment.

\begin{figure}[!t]
  \centering
  
  \includegraphics[width=1\linewidth]{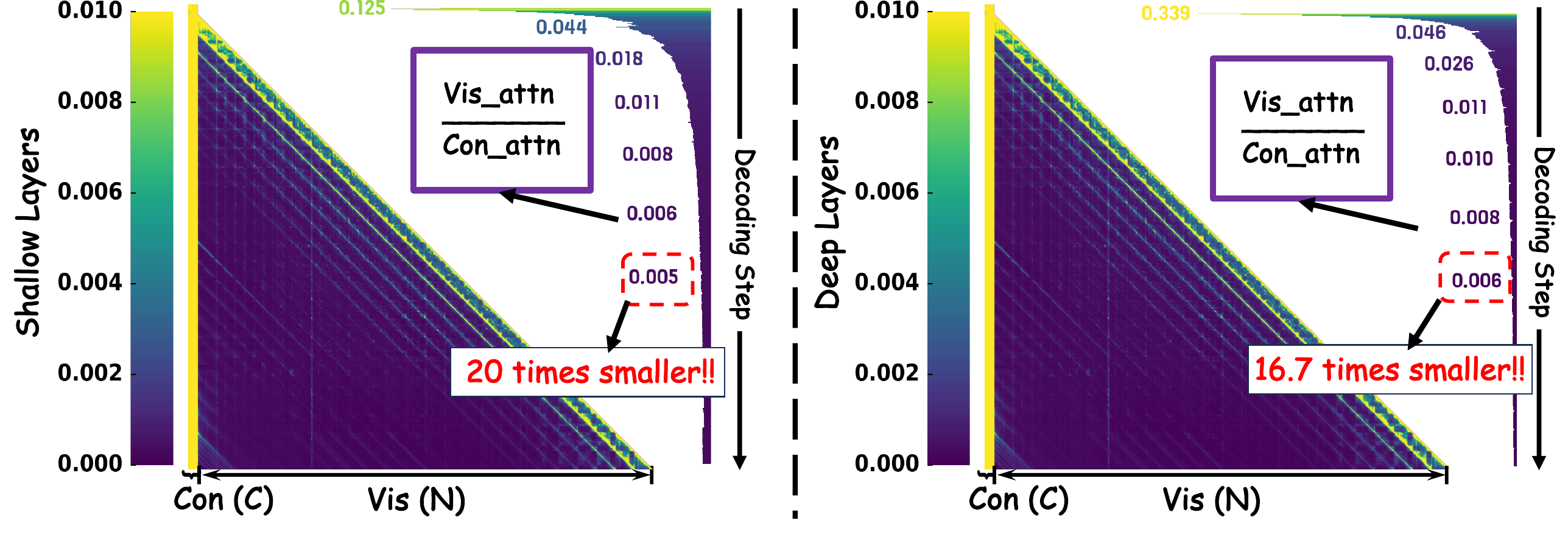}
    \caption{Visualization of attention patterns and allocation. The attention persistently shifts toward the conditional tokens, while visual attention gradually dilutes as decoding progresses.}
  \label{ovser1}
  \vspace{-0.25cm}
\end{figure}
\begin{figure}[!t]
  \centering
  
  \includegraphics[width=1\linewidth]{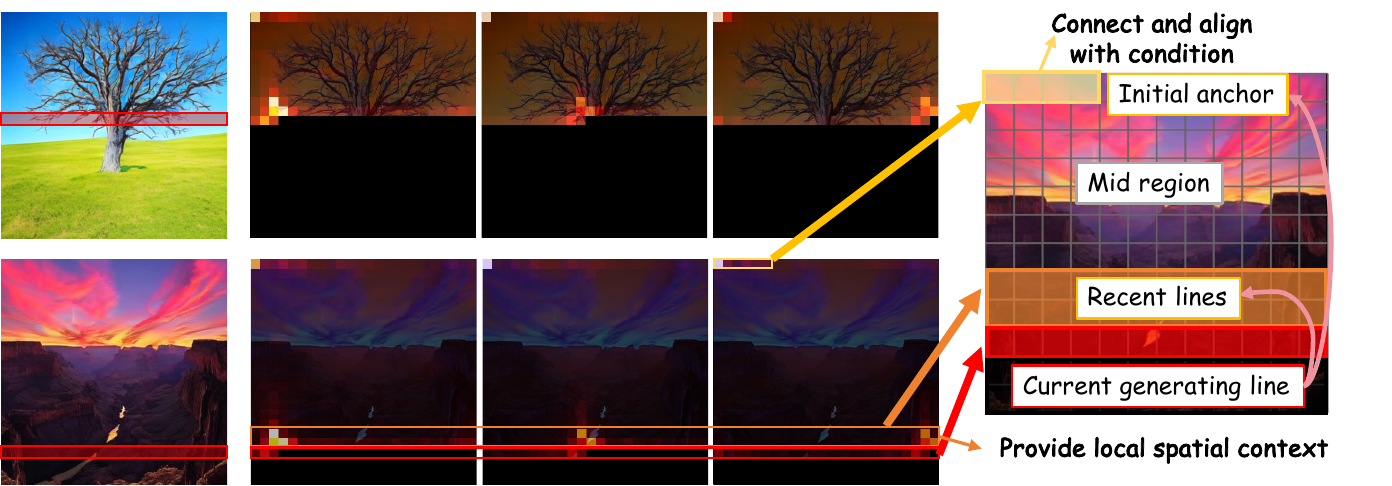}
    \caption{Visualization of attention evolution in line generation. The current line generation relies on tokens from the recent region and initial anchor, resulting in mid-region cache redundancy.}
  \label{ovser2}
   \vspace{-0.28cm}
\end{figure}

\textit{Is caching all visual tokens truly necessary?} To address this question, we revisit autoregressive image generation models and identify substantial inefficiencies in their attention mechanisms.
As shown in Figure\,\ref{ovser1}, attention gradually shifts towards conditional text tokens, while the attention received by each historical visual token decreases throughout the sequential decoding process. We find that this inefficiency in visual attention arises from its locally concentrated visual dependency.
We visualize the evolution of attention during image generation from a 2D perspective. As shown in Figure\,\ref{ovser2}, each generated token predominantly attends to its spatially adjacent tokens that provide nearby spatial context, and the initial visual anchor that aligns with the conditional tokens. Consequently, the generation of each image line primarily relies on tokens from recent lines and the initial visual anchor, while distant tokens from earlier lines contribute minimally to subsequent generations.
This suggests that maintaining all historical visual tokens in the cache is inefficient, and the visual KV cache should be dynamically compressed during decoding.

\begin{figure}[!t]
  \centering
  \includegraphics[width=1.0\linewidth]{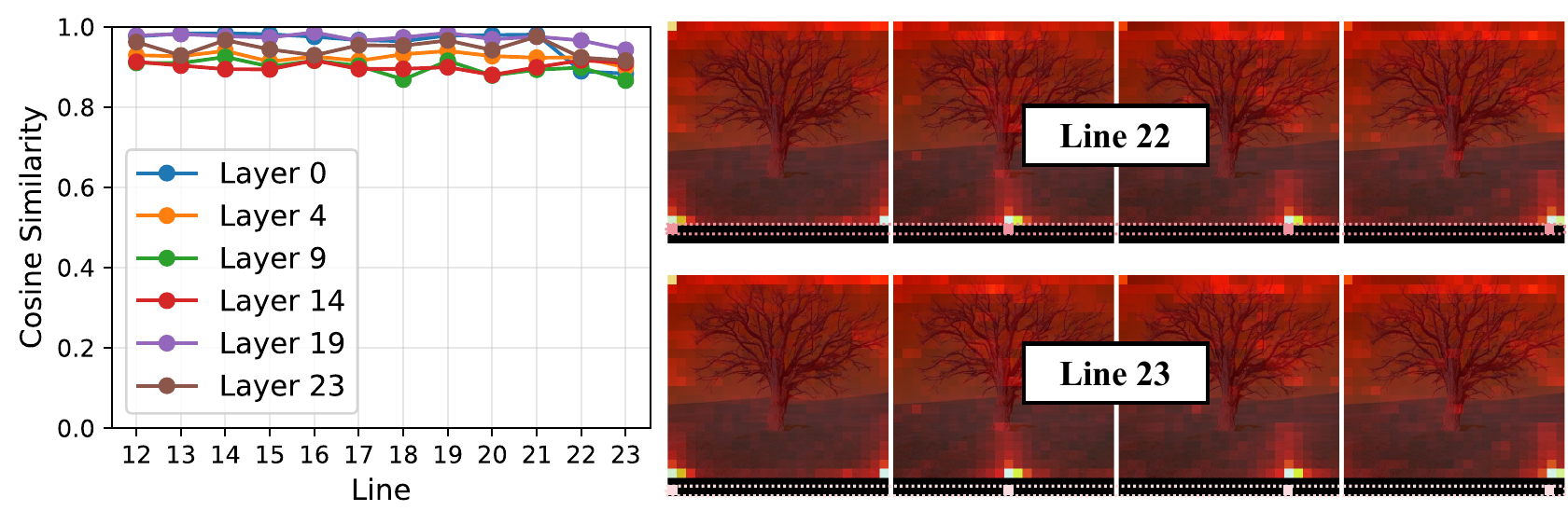}
\caption{(Left) Attention similarity between adjacent lines. (Right) Attention visualization for past generated regions across two adjacent lines. Inter-line attention shows high consistency.}
  \label{obser3}
\end{figure}

\textit{How can we improve KV cache compression for autoregressive image generation models?}
Building on the inefficiencies of vanilla visual caching, we explore KV cache compression to reduce redundant storage and computation during AR image generation.
Existing compression methods are typically designed for the prefilling phase of textual contexts in LLMs, often using coarse, one-shot truncation strategies. However, AR visual models progressively build a large cache during decoding, where token importance evolves dynamically with spatial progression. This calls for an adaptive and progressive compression pipeline, where each compression step should not hinder later generation. Additionally, unlike text, which follows a one-dimensional sequence with varying relevance across adjacent sentences, images are two-dimensional and exhibit strong locality and continuity across neighboring regions. As shown in Figure\,\ref{obser3}, adjacent raster lines often share similar attention patterns, revealing strong inter-line consistency that could provide useful contextual cues for generating subsequent lines. These stage-wise and modality-specific differences highlight the gap in decoding-time KV cache compression research for AR visual models.


Building on these insights, we propose LineAR, a novel training-free progressive KV cache compression pipeline tailored for autoregressive visual models. Specifically, LineAR manages the KV cache from a 2D view by dividing the image generation process into rasterized line stages. This enables efficient cache management that fully exploits the locality and inter-line consistency inherent in visual generation. By maintaining initial anchor tokens and recent lines to preserve global conditioning and local dependencies, LineAR progressively removes less informative tokens for next line generation under inter-line guidance. As a result, LineAR frees AR visual models from heavy KV cache dependency, allowing image generation with only a few lines of cached tokens.
LineAR achieves significant memory reduction and throughput acceleration while maintaining or even improving generation quality, all without requiring any retraining.
The contributions of our paper include:
\begin{itemize}
    \item We conduct an in-depth analysis of AR visual models, revealing inefficiencies in visual attention during the decoding process and identifying inherent visual characteristics, including locally concentrated visual dependencies and inter-line attention consistency.
    
    \item We propose a simple yet effective decoding-time KV cache compression pipeline that performs progressive compression under inter-line guidance, enabling AR visual models to operate with only a few lines of cache.
    
    \item We extensively validate LineAR on six AR visual models across various benchmarks and evaluation metrics. Experiments demonstrate that LineAR achieves significant memory savings and throughput acceleration while maintaining lossless or even improved generation quality.
\end{itemize}

\section{Related Work}
\label{sec:related}

\textbf{Autoregressive Visual Generation.}
Autoregressive (AR) models~\cite{achiam2023gpt,anthropic2024claude3,dubey2024llama,liu2024deepseek, yang2025qwen3} have shown great success in text generation, leading to their extension into image synthesis.
Pioneering works~\cite{ramesh2021zero,sun2024autoregressive,he2024mars,wang2025simplear,yue2025understand} paved the way by quantizing images into discrete tokens~\cite{van2017neural,yu2021vector,esser2021taming}, decoded autoregressively in a GPT-style manner.
Building on this, recent advances~\cite{wang2024emu3,liu2024lumina-mgpt,xin2025lumina,chen2025janus,wu2024janus,chameleonteam2025chameleonmixedmodalearlyfusionfoundation} adopt architectures closely aligned with text generation, enabling a unified framework for multimodal understanding and generation. 
However, next-token prediction suffers from severe inference inefficiency, as each token must be generated sequentially. 
To alleviate this, alternative designs explore more efficient paradigms, such as next-scale~\cite{tian2024visual,han2024infinity,tang2024hart,ren2024m,voronov2024switti} and masked autoregressive modeling~\cite{chang2022maskgit, li2022mage, li2024autoregressive}. 
Despite these, all methods still require caching previously generated tokens, causing significant storage overhead. Our work focuses on the next-token paradigm, the most general and widely adopted across modalities.

\textbf{Efficient Autoregressive Visual Generation.}
Recent efforts have focused on improving the efficiency of visual AR generation.
Parallel decoding~\cite{teng2024accelerating, zhang2025locality,he2025neighboring,hezipar,huang2025spectralar,pang2025randar,wang2025parallelized} approaches aim to reduce decoding steps by predicting multiple visual tokens simultaneously, thereby accelerating generation.
Despite their effectiveness, they often require retraining AR models to learn alternative decoding orders~\cite{wang2025parallelized,pang2025randar,he2025neighboring,zhang2025locality}, or trade additional computation~\cite{teng2024accelerating,hezipar} for reduced latency.
More importantly, they fail to reduce the memory footprint of cached tokens during generation, resulting in nearly unchanged peak memory usage. Other strategies, such as token sparification~\cite{guo2025fastvar, chen2025sparsevar, qin2025head} and cache reuse~\cite{yan2025lazymar}, optimize generation efficiency but are typically tailored to specific paradigms, making them unsuitable for the next-token prediction approach.

\textbf{KV Cache Compression.}
KV cache compression~\cite{zhang2024h2o,liu2024kivi,kang2024gear,wan2024look} has emerged as a promising solution to address the memory bottleneck in autoregressive models. Eviction-based methods~\cite{zhang2024h2o, liu2024scissorhands,oren2024transformers,ren2024efficacy,li2024snapkv,ge2023model,yang2024pyramidinfer,feng2024ada,fu2024not,qin2025cake} effectively control cache size by enforcing a target budget. 
While these techniques show strong results in LLMs, they mostly focus on prefill phase KV compression, truncating redundant tokens from long inputs in a one-shot manner.
In contrast, autoregressive visual models generate long token sequences progressively during decoding, 
where token importance evolves dynamically. 
Although methods like StreamingLLM~\cite{xiao2023efficient} and H2O~\cite{zhang2024h2o} enable decoding-time KV cache compression, and recent work like R-KV~\cite{cai2025r} optimizes reasoning-oriented LLMs~\cite{guo2025deepseek} by considering contextual redundancy, these approaches fail to address the spatial redundancy and the need for spatially consistent cache management in visual generation, making them suboptimal for visual AR models.
To fill this gap, our work pioneers a dedicated decoding-time KV cache compression pipeline tailored for AR image generation.

\begin{figure}[t]
  \centering
  
  \includegraphics[width=1\linewidth]{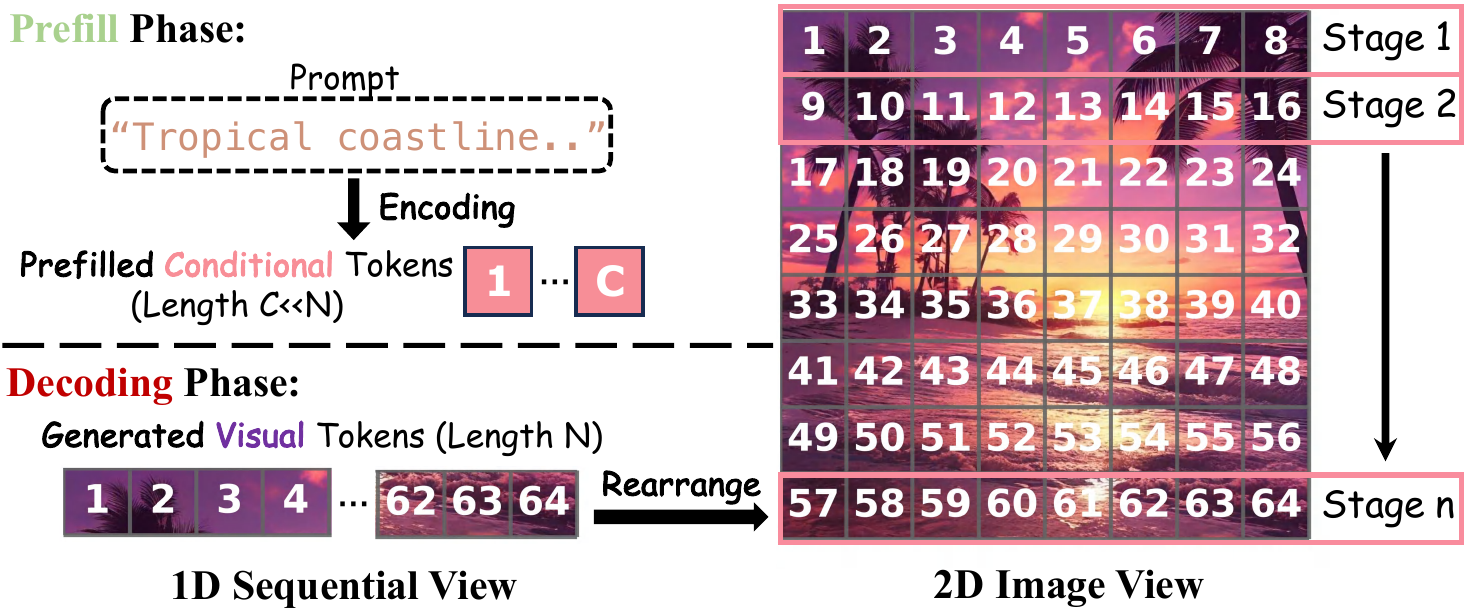}
    \caption{Visual KV cache management from a 2D perspective.
    The visual KV cache is organized in a 2D view, where each horizontal line represents a generating stage and serves as a natural unit for cache management in LineAR.}
  \label{fig:2dview}
\end{figure}
\section{Method}
\label{method}
\subsection{Preliminaries}

\textbf{Autoregressive Visual Generation.}
Autoregressive (AR) visual generation follows next-token paradigm to synthesize visual content in raster order. Given a textual prompt (or class label) tokenized into a sequence of condition tokens $\mathbf{c}\in\mathbb{R}^{C}$, the model generates a sequence of visual tokens $(\mathbf{x}_1,\mathbf{x}_2,\ldots,\mathbf{x}_N)$, where each token is predicted based on the conditional prefix and the previously generated tokens:
\begin{equation}
\small
p(\mathbf{x}_{1:N}\mid \mathbf{c})=\prod_{i=1}^{N} p(\mathbf{x}_i \mid \mathbf{c}, \mathbf{x}_{<i}),
\end{equation}
where $N=h\times w$ denotes the number of visual tokens on the latent feature grid. After decoding, the tokens are mapped back to pixel space by a pretrained visual decoder and rearranged to form an image of size $H\times W$.





\begin{figure*}[t]
  \centering
  
  \includegraphics[width=1\linewidth]{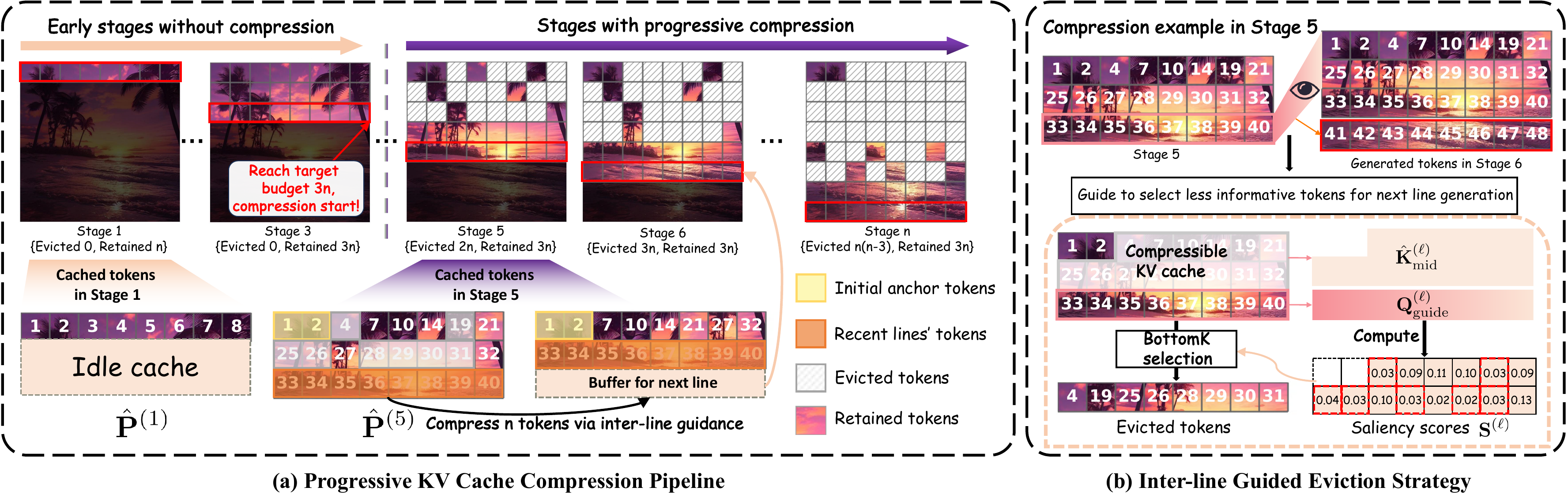}
    \caption{Overview of the proposed LineAR framework. We illustrate a toy example of $8 \times 8$ image grid with a budget ratio $\rho=3/8$ in one attention head, showing (a) the Progressive KV Cache Compression Pipeline, which progressively reduces cached visual tokens, and (b) the Inter-line Guided Eviction Strategy, which evicts less informative tokens based on inter-line attention guidance.}
  \label{fig:overall}
\end{figure*}
\textbf{KV Cache in Visual Generation.}
Autoregressive (AR) visual generation typically involves two phases: (1) \emph{prefill phase}: the model encodes text prompts or class labels into $C$ conditional tokens, forming the conditional key–value (KV) states 
$\{\mathbf{K}_{\text{cond}}, \mathbf{V}_{\text{cond}}\} \in \mathbb{R}^{C\times d}$.
(2) \emph{decoding phase}: the model autoregressively generates subsequent visual tokens in multiple decoding steps.
For a single attention head at decoding step $i$, the attention is computed as:
\begin{equation}
\small
\mathrm{Attention}(\mathbf{q}_{i}, \mathbf{K}^{(i)}, \mathbf{V}^{(i)})
= \mathrm{Softmax}\!\left(\frac{\mathbf{q}_{i}{\mathbf{K}^{(i)}}^{\top}}{\sqrt{d}}\right)\mathbf{V}^{(i)},
\end{equation}
where $\mathbf{q}_{i}\!\in\!\mathbb{R}^{1\times d}$ is the query of the $i$-th visual token, and 
$\{\mathbf{K}^{(i)}, \mathbf{V}^{(i)}\}\!\in\!\mathbb{R}^{(C+i)\times d}$ denote the current KV states consisting of both conditional and visual tokens:
\begin{equation}
\small
\mathbf{K}^{(i)} = [\mathbf{K}_{\text{cond}},\,\mathbf{K}_{1:i}], 
\qquad 
\mathbf{V}^{(i)} = [\mathbf{V}_{\text{cond}},\,\mathbf{V}_{1:i}].
\end{equation}

During inference, these KV pairs are cached to avoid recomputing projections for previous tokens. 
At any decoding step $i$, the visual KV cache is updated by concatenating the previous cache 
$\{\mathbf{K}_{1:i-1}, \mathbf{V}_{1:i-1}\}$ 
with the newly generated KV pair $\{\mathbf{k}_{i}, \mathbf{v}_{i}\}$, forming:
\begin{equation}
\small
\label{kvcat}
\mathbf{P}_{1:i} 
= \big\{\mathbf{K}_{1:i}, \mathbf{V}_{1:i}\big\}
= \big\{ [\mathbf{K}_{1:i-1}, \mathbf{k}_{i}],\, [\mathbf{V}_{1:i-1}, \mathbf{v}_{i}] \big\}.
\end{equation}

For simplicity, we denote the visual KV cache at step $i$ by $\mathbf{P}_{1:i}$.
While KV cache effectively eliminates redundant computation, its memory footprint still grows linearly with the number of generated visual tokens $N$, 
resulting in $\mathcal{O}(N)$ storage that rapidly dominates inference cost at high resolutions and severely limits decoding efficiency.

\subsection{Motivation}
The growing KV cache size creates a bottleneck in AR image generation, motivating us to optimize cache usage by controlling its expansion during decoding. To address this, we analyze the autoregressive attention mechanism and ask two key questions: (1) Is it necessary to cache all visual tokens? (2) If not, how can we more efficiently handle past visual tokens? We perform an in-depth analysis using Janus-Pro~\cite{chen2025janus} on 300 prompts from the DPG dataset~\cite{hu2024ella}, which guides the design of our proposed method.

\textit{\textbf{Observation 1: Inefficient Visual Attention in AR Models}.} Figure\,\ref{ovser1} illustrates the attention patterns and the attention allocation ratio between visual and conditional tokens in autoregressive visual generation.
At early decoding steps, visual tokens closely align with conditional tokens to establish the overall visual style and color tone of the generated image~\cite{xiang2025makeefficientdynamicsparse}, thereby receiving a comparable attention allocation. However, as generation proceeds, visual attention becomes increasingly diluted, while attention persistently sinks toward the conditional tokens, whose average attention allocation even grows up to be $20\times$ higher than that of visual tokens in shalldow layrers.
In models such as Lumina-mGPT~\cite{liu2024lumina-mgpt}, the number of cached visual tokens can exceed the conditional ones by more than 200$\times$, making the vanilla KV cache mechanism highly inefficient for autoregressive visual generation.

\textit{\textbf{Observation 2: Locally Concentrated Visual Dependency}.} 
To further understand how visual attention evolves during decoding, we observe the generation of image lines, which aligns with the raster-scan order in autoregressive visual models. As shown in Figure\,\ref{obser3}, attention is primarily focused on spatially adjacent tokens and the initial anchor tokens that are strongly linked to the conditional tokens. This highlights that visual generation relies heavily on local spatial context and persistent dependence on the initial anchor tokens that encode the global style and conditional alignment.
Consequently, the generation of the current line is mainly dependent on neighboring lines and initial anchor tokens, while distant visual tokens contribute little to the prediction of subsequent tokens. This localized attention distribution reveals the redundancy of distant visual tokens, making it intuitive and feasible to progressively discard outdated tokens during the generation process.

\textit{\textbf{Observation 3: Inter-Line Attention Consistency}}. To progressively compress visual tokens during decoding, it is crucial to ensure that each compression step does not disrupt the generation of subsequent tokens.
Leveraging the spatial continuity in images, we hypothesize that adjacent lines attend to highly similar visual regions during generation. To verify this, we compare the attention patterns of consecutive lines in the later stages of decoding, focusing on the previously generated content. As shown in Figure\,\ref{obser3}, the attention of adjacent lines exhibits strong overlapping, with the similarity curve further quantifying the degree of consistency across layers and lines.
This indicates that the information attended to by the generation process in adjacent lines is highly similar. Consequently, the attended regions from adjacent line can serve as reliable guidance for determining which tokens should be retained for the next-line generation, enabling safe and adaptive compression.

\subsection{LineAR}
We propose \textit{LineAR}, a novel KV cache compression pipeline that enables high-quality autoregressive image generation with only a few lines of cached tokens.
 

\textbf{Visual KV Cache Management in a 2D View.}
In line with our analysis, we manage the visual KV cache from a \emph{2D view}, which better reflects image structure than the vanilla sequential view.
Let the total number of visual tokens be $N = h \times w = n^2$ for simplicity. 
As illustrated in Figure\,\ref{fig:2dview}, the full KV cache can be viewed as an $n \times n$ matrix that follows raster order, where each position corresponds to a latent image feature.
Under this view, the generation process can be naturally regarded as a sequence of $n$ raster-line stages, 
each corresponding to the generation of one horizontal line consisting of $n$ visual tokens.
We denote by 
$ \mathbf{P}^{(l)}\triangleq
\{\mathbf{K}^{(l)} = \mathbf{K}_{1:nl},  
\mathbf{V}^{(l)} = \mathbf{V}_{1:nl}\}
$
the state of the KV cache after the $l$-th line has been generated. 
This 2D view introduces a natural management unit, the raster line, which serves as the fundamental element for cache organization, allowing the model to maintain spatial coherence during progressive compression.

\textbf{Cache Budget and Budget Ratio.}
We set the cache budget $B$ as the upper bound on the number of cached KV pairs and assume that it is an integer multiple of $n$, \emph{i.e.}, $n \mid B$, so that the cache can be managed at the granularity of complete raster lines. 
We further define the budget ratio $\rho = B / N$.

\textbf{Progressive KV Cache Compression Pipeline.} 
Figure\,\ref{fig:overall}(a) illustrates the overall pipeline. 
Given a target budget ratio $\rho$, we maintain a cache space upper bounded by $|\hat{\mathbf{P}}|\le\rho N=B$ during generation.
The cache is regulated in a stable and orderly manner, using the end of each raster line as the synchronization point. 
Recall that we split AR image generation into $n$ lines, in the early lines' generation ($\ell/n<\rho$), no compression is applied, allowing the model to first establish a consistent global style and remain fully guided by the conditional context. Once the generated context exceeds the predefined cache bound, progressive KV compression is activated in the following lines. 

At any end of line $\ell$ when compression starts, $\hat{\mathbf{P}}^{(\ell)}$ can be always partitioned into three parts: 
(1) Initial anchor tokens $\hat{\mathbf{P}}^{(\ell)}_{\text{init}}=\mathbf{P}_{1{:}N_{\text{init}}}$ that maintain the global style and conditional alignment;
(2) Recent lines' tokens $\hat{\mathbf{P}}^{(\ell)}_{\text{rec}}=\mathbf{P}_{(\ell{-}r)n+1:\ell n}$ that provide local visual context; and (3) The mid-region tokens $\hat{\mathbf{P}}^{(\ell)}_{\text{mid}}$, which lie between the initial and recent windows and are compressibe in the current stage.
For $\hat{\mathbf{P}}^{(\ell)}_{\text{mid}}$, we then perform the inter-line guided eviction strategy $\Phi_{\text{inter}}(\cdot)$, which evicts the least $n$ informative tokens for next line generation, $\hat{\mathbf{P}}^{(\ell)}_{\text{evict}} \in \mathbb{R}^{n \times d}$. We compact mid-region cache $\hat{\mathbf{P}}^{(\ell)}_{\text{mid}}$ by removing the evicted tokens:
\begin{equation}
\small
\hat{\mathbf{P}}^{(\ell)}_{\text{evict}} = \Phi_{\text{inter}}(\hat{\mathbf{P}}^{(\ell)}_{\text{mid}}, n),
\quad \hat{\mathbf{P}}^{(\ell)}_{\text{mid}} \leftarrow \hat{\mathbf{P}}^{(\ell)}_{\text{mid}} \setminus \hat{\mathbf{P}}^{(\ell)}_{\text{evict}}.
\end{equation}
After compression, the cache $\hat{\mathbf{P}}^{(\ell)}$ is updated by:
\begin{equation}
\small
\hat{\mathbf{P}}^{(\ell)} \leftarrow \big[\hat{\mathbf{P}}^{(\ell)}_{\text{init}},\hat{\mathbf{P}}^{(\ell)}_{\text{mid}},\hat{\mathbf{P}}^{(\ell)}_{\text{rec}}\big],\quad |\hat{\mathbf{P}}^{(\ell)}|=B-n.
\label{eq:compact_pipeline}
\end{equation}

Note that, at each compression operation, we preserve an $n$-token buffer for the next line generation. So at the next stage $\ell{+}1$, the cache is first updated by concatenating the new generated tokens for line $\ell{+}1$: 
\begin{equation} \hat{\mathbf{P}}^{(\ell+1)} =\big[\hat{\mathbf{P}}^{(\ell)},\ \mathbf{P}_{\ell n+1:(\ell+1)n}\big], \end{equation} 
and then the compression operation is repeated iteratively. This completes the progressive KV cache compression pipeline, maintaining anchors and locality, enforcing the budget, and leaving a one-line buffer, thus balancing compression frequency and generation stability.

\textbf{Inter-line Guided Eviction Strategy $\Phi_{\text{inter}}$.}
In the progressive compression pipeline, the initial anchor $\hat{\mathbf{P}}^{(\ell)}_{\text{init}}$ is fixed and the recent window $\hat{\mathbf{P}}^{(\ell)}_{\text{rec}}$ advances line by line.
The key challenge is to prudently evict tokens from the mid-region without affecting the generation of the next line. Leveraging the strong visual consistency between adjacent image lines, we design the inter-line guided eviction strategy (shown in Figure\,\ref{fig:overall}(b)), which utilizes similarity information from the adjacent line to guide token selection.

For the uncompressed key-value states of the remaining mid tokens $\hat{\mathbf{P}}^{(\ell)}_{\text{mid}}=\{\hat{\mathbf{K}}^{(\ell)}_{\text{mid}},\hat{\mathbf{V}}^{(\ell)}_{\text{mid}}\}$, our goal is to evict $n$ tokens that are least informative for the next line's generation. 
To achieve this, we always maintain a guide queue of queries from the current generating line $\ell$:
\begin{equation}
\small
\mathbf{Q}^{(\ell)}_{\text{guide}} = [\,\mathbf{q}_{(\ell-1)n+1},\ldots,\mathbf{q}_{\ell n}\,]\in\mathbb{R}^{n\times d}.
\end{equation}

We compute saliency scores for the mid-region by averaging the attention from the current line’s queries:
\begin{equation}
\small
\mathbf{S}^{(\ell)} = 
\frac{1}{n}\,\mathbf{1}^{\top}\mathrm{Softmax}\!\Big(
\frac{\mathbf{Q}^{(\ell)}_{\text{guide}}\,
\hat{\mathbf{K}}^{(\ell)\top}_{\text{mid}}}{\sqrt{d}}
\Big)\in\mathbb{R}^{|\hat{\mathbf{P}}^{(\ell)}_{\text{mid}}|},
\label{eq:score}
\end{equation}
which ensures that each token from the current line equally contributes to guiding the eviction decision.
We then select the $n$ least salient tokens to be evicted:
\begin{equation}
\small
\mathcal{E}^{(\ell)}=\mathrm{BottomK}\!\big(\mathbf{S}^{(\ell)},\,n\big),\quad \hat{\mathbf{P}}^{(\ell)}_{\text{evict}}
= \hat{\mathbf{P}}^{(\ell)}_{\text{mid}}[\mathcal{E}^{(\ell)}].
\end{equation}

In this way, we preserve the most important information, achieving lossless compression for the next line generation.

\begin{table*}[t]
\footnotesize
\caption{Quantitative results of text-to-image generation on Janus-Pro-1B, Janus-Pro-7B, Lumina-mGPT-768 and Lumina-mGPT-1024 with LineAR under different budget ratios. We compare the speedup with the full-cache maximum throughput, using the same batch size.}
\label{t2iresults}
\centering
\setlength{\tabcolsep}{4pt}

\begin{tabular}{c c ccccc cccccc}
\toprule
\multirow{2}{*}{\textbf{Models}} &
\multirow{2}{*}{\textbf{$\rho$}}  &
\multicolumn{2}{c}{\textbf{Efficiency}$\uparrow$} &
\multicolumn{5}{c}{\textbf{GenEval}$\uparrow$}&\multicolumn{4}{c}{\textbf{DPG}$\uparrow$} \\
\cmidrule(lr){3-4}\cmidrule(lr){5-9} \cmidrule(lr){10-13}
&    & Throughput &Speedup& Single Obj. & Two Obj. & Counting & Colors & Overall &Entity & Relation & Attribute & Overall\\
\midrule
\multirow{3}{*}{\thead{Janus-Pro-1B\\($N=576$)}} & Full &0.89$\text{it/s}$ &1.00$\times$  &0.98 & 0.83 &0.50 &0.90 &0.73 &90.40 &92.80 &88.78 &85.00 \\
  & \cellcolor{gray!20}1/4   &\cellcolor{gray!20}4.03$\text{it/s}$ &\cellcolor{gray!20}4.53$\times$ & \cellcolor{gray!20}0.99&\cellcolor{gray!20}0.86  &\cellcolor{gray!20}0.49  & \cellcolor{gray!20}0.89 & \cellcolor{gray!20}0.72 & \cellcolor{gray!20}90.26 & \cellcolor{gray!20}93.19   & \cellcolor{gray!20}88.26 &  \cellcolor{gray!20}84.79\\
  & \cellcolor{gray!20}1/6   &\cellcolor{gray!20}5.22$\text{it/s}$ &\cellcolor{gray!20}5.87$\times$ & \cellcolor{gray!20}0.99& \cellcolor{gray!20}0.81 &\cellcolor{gray!20}0.51  &\cellcolor{gray!20}0.89  &  \cellcolor{gray!20}0.71& \cellcolor{gray!20}89.92  & \cellcolor{gray!20}92.57 & \cellcolor{gray!20}88.26 & \cellcolor{gray!20}84.59 \\
\midrule
\multirow{3}{*}{\thead{Janus-Pro-7B\\($N=576$)}}  & Full&0.35$\text{it/s}$ &1.00$\times$  &0.99 & 0.86 &0.57& 0.91 &0.79 &90.34 &93.23 &88.28 &85.54\\
  &\cellcolor{gray!20}1/4   &\cellcolor{gray!20}1.58$\text{it/s}$ &\cellcolor{gray!20}4.49$\times$  &\cellcolor{gray!20}0.99  &\cellcolor{gray!20}0.89  &\cellcolor{gray!20}0.58 &\cellcolor{gray!20}0.93& \cellcolor{gray!20}0.80& \cellcolor{gray!20}90.52 &\cellcolor{gray!20}93.50  & \cellcolor{gray!20}88.22 & \cellcolor{gray!20}85.60 \\
  &\cellcolor{gray!20}1/6   &\cellcolor{gray!20}1.98$\text{it/s}$  &\cellcolor{gray!20}5.62$\times$ & \cellcolor{gray!20}0.99 &\cellcolor{gray!20}0.88 &\cellcolor{gray!20}0.52 &\cellcolor{gray!20}0.91 &\cellcolor{gray!20}0.78& \cellcolor{gray!20}90.32 & \cellcolor{gray!20}93.38 & \cellcolor{gray!20}87.97&\cellcolor{gray!20}85.30 \\

\midrule

\multirow{3}{*}{\thead{Lumina-mGPT-768 \\($N=2352$)}}& Full &0.04$\text{it/s}$ &1.00$\times$ &1.00  &0.73 &0.25 &0.82 &0.52  &83.54 &87.81 &81.57 &76.96 \\
  & \cellcolor{gray!20}1/6   &\cellcolor{gray!20}0.08$\text{it/s}$ &\cellcolor{gray!20}1.93$\times$ &\cellcolor{gray!20}0.99 &\cellcolor{gray!20}0.74 &\cellcolor{gray!20}0.28 &\cellcolor{gray!20}0.84 &\cellcolor{gray!20}0.52  &\cellcolor{gray!20}84.46 &\cellcolor{gray!20}88.39 &\cellcolor{gray!20}82.01 &\cellcolor{gray!20}77.76 \\
  &\cellcolor{gray!20}1/8   &\cellcolor{gray!20}0.09$\text{it/s}$ &\cellcolor{gray!20}2.13$\times$ &\cellcolor{gray!20}0.99 &\cellcolor{gray!20}0.73 &\cellcolor{gray!20}0.24 &\cellcolor{gray!20}0.86 &\cellcolor{gray!20}0.52  &\cellcolor{gray!20}83.89 &\cellcolor{gray!20}87.89 &\cellcolor{gray!20}82.29 &\cellcolor{gray!20}77.19 \\

\midrule

\multirow{2}{*}{\thead{Lumina-mGPT-1024 \\($N=4160$)}} & Full  &0.01$\text{it/s}$ &1.00$\times$  & 0.98 & 0.76 & 0.29& 0.83 & 0.56 &86.67 &90.95  &84.42 &79.98 \\
   & \cellcolor{gray!20}1/8  &\cellcolor{gray!20}0.07$\text{it/s}$ &\cellcolor{gray!20}2.64$\times$ & \cellcolor{gray!20}0.99 & \cellcolor{gray!20}0.75 & \cellcolor{gray!20}0.30 & \cellcolor{gray!20}0.82 &\cellcolor{gray!20}0.55  &\cellcolor{gray!20}86.67  &\cellcolor{gray!20}90.91 &\cellcolor{gray!20}84.22 &\cellcolor{gray!20}79.65 \\

\bottomrule
\end{tabular}
\vspace{-0.1cm}
\end{table*}

\section{Experimentation}
\subsection{Experimental Setup}

\textbf{Evaluated Models.} We evaluate LineAR on six state-of-the-art autoregressive image generation models, covering three representative architectures.
(1) Janus-Pro-1B/7B~\cite{chen2025janus}, unified multimodal models generating $384 \times 384$ images with 576 visual tokens;
(2) Lumina-mGPT-768/1024~\cite{liu2024lumina-mgpt}, text-to-image models generating $768 \times 768$ and $1024 \times 1024$ images, with 2352 and 4160 visual tokens;
(3) LlamaGen-XL/XXL~\cite{sun2024autoregressive}~\cite{sun2024autoregressive}, class-conditional models generating $384 \times 384$ images with 576 visual tokens.

\textbf{Evaluation Metrics.} 
For text-to-image generation, we adopt two widely used benchmarks, GenEval~\cite{ghosh2023geneval} and DPG~\cite{hu2024ella}, to assess high-level semantic alignment and compositional consistency.  
We further evaluate image quality and prompt adherence using Fréchet Inception Distance~\cite{heusel2017gans} (FID) and CLIP score~\cite{radford2021learning} on 30k MS-COCO~\cite{lin2014microsoft} validation captions. We denote FID for MS-COCO as FID-30k.
For class-conditional image generation, we follow standard ImageNet evaluation protocols and report FID-50k, Inception Score (IS), Precision, and Recall, computed from 50k generated samples.

\textbf{Implementation Details.}
We report results of LineAR under different budget ratio settings $(\rho = B / N)$. For each evaluated model, we follow its original inference configuration. Additional implementation details of LineAR for each model are provided in the \textit{supplementary material}. Our experiments are conducted on A800 GPUs.
\begin{figure*}[t]
  \centering
  
  \includegraphics[width=1\linewidth]{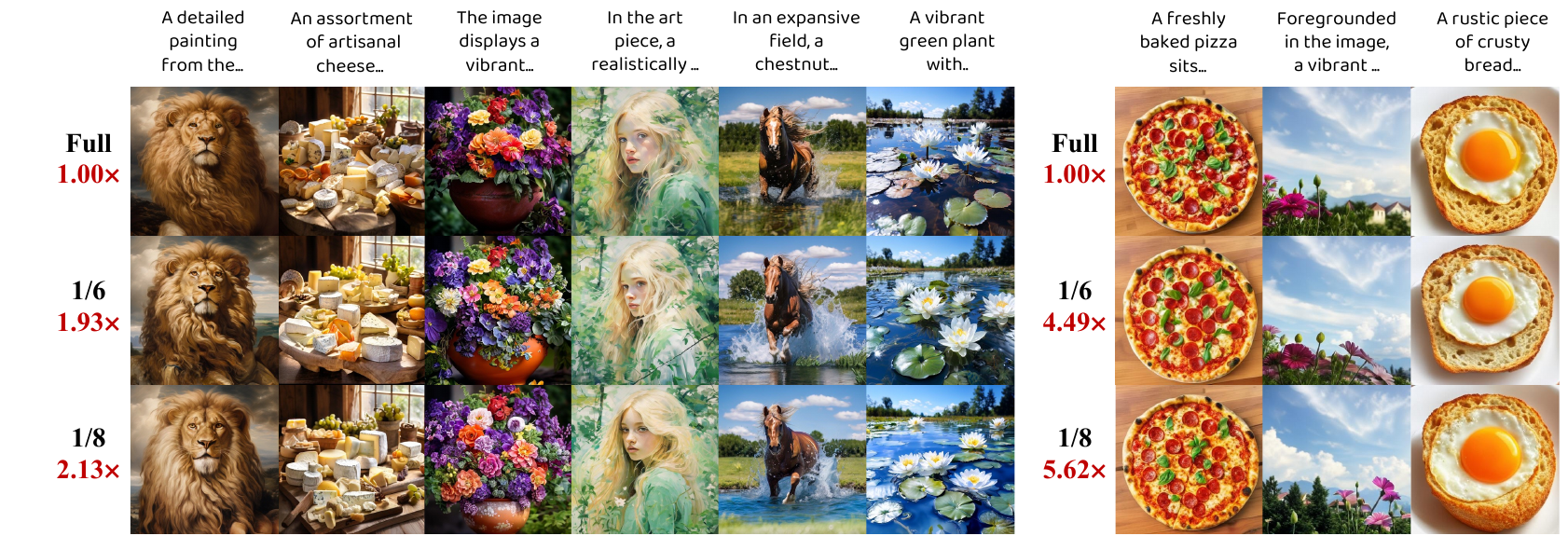}
    \caption{Qualitative results of text-to-image generation by LineAR on Lumina-mGPT-768 (left) and Janus-Pro-7B (right).}
  \label{quant}
\vspace{-0.25cm}
\end{figure*}

\subsection{Main Results}
\textbf{Text-to-Image Generation.}
\textit{Quantitative results.}
Table\,\ref{t2iresults} presents the results of applying LineAR to four models spanning different architectures, model sizes, and generation resolutions. 
Across all cases, LineAR consistently achieves lossless or even improved performance compared to the full-cache baseline. 
For instance, on Lumina-mGPT-768, LineAR achieves higher DPG scores with only 1/8 visual cache retained. 
Beyond its strong performance on high-resolution generation, LineAR also performs effectively on Janus-Pro models, which generate low-resolution images with only 576 visual tokens. 
Even when the cache is reduced to 1/6, LineAR maintains nearly identical quality, exhibiting only minor drops of 0.02 and 0.01 in GenEval and 0.41 and 0.24 in DPG compared to the full-cache baseline, while achieving 5.87$\times$ and 5.62$\times$ throughput speedup on Janus-Pro-1B/7B, respectively.

\textit{Qualitative results.}
Figure\,\ref{quant} presents images generated with Lumina-mGPT-768 and Janus-Pro-7B using LineAR under different cache ratios. Across models, LineAR preserves high visual quality that remains well aligned with textual prompts, maintaining similar style and color tones to the full cache, even under high compression ratios. This demonstrates the effectiveness of LineAR, designed with the characteristics of visual generation in mind.

\begin{table}[t]
\footnotesize
\caption{Quantitative results of class-to-image generation on LlamaGen-XL/XXL with LineAR under different budget ratios.}
\label{llamagenresults}
\centering
\setlength{\tabcolsep}{4pt}
\begin{tabular}{cc cccc}
\toprule
\textbf{Models} &
\textbf{$\rho$}  &
\textbf{FID-50k}$\downarrow$ & 
\textbf{IS}$\uparrow$ & \textbf{Precision}$\uparrow$ &\textbf{Recall$\uparrow$}\\
\midrule
\multirow{4}{*}{\thead{LlamaGen-XL \\($N=576$)}} & Full  & 2.77  & 282.41 &0.83  &0.54\\
   & \cellcolor{gray!20}1/4   & \cellcolor{gray!20}2.64  & \cellcolor{gray!20}279.95&\cellcolor{gray!20}0.83  & \cellcolor{gray!20}0.55 \\
   & \cellcolor{gray!20}1/6  & \cellcolor{gray!20}2.68   & \cellcolor{gray!20}279.17&\cellcolor{gray!20}0.82  & \cellcolor{gray!20}0.55\\ 
   &\cellcolor{gray!20}1/8  & \cellcolor{gray!20}2.95  &\cellcolor{gray!20}282.91&\cellcolor{gray!20}0.82  &\cellcolor{gray!20}0.54 \\
   \midrule
\multirow{4}{*}{\thead{LlamaGen-XXL \\($N=576$)}} & Full  & 2.47 & 290.82 &0.83  &0.56\\
& \cellcolor{gray!20}1/4   & \cellcolor{gray!20}2.45 & \cellcolor{gray!20}292.47& \cellcolor{gray!20}0.83  & \cellcolor{gray!20}0.56 \\
& \cellcolor{gray!20}1/6  & \cellcolor{gray!20}2.54 &\cellcolor{gray!20}290.14&\cellcolor{gray!20}0.82  &\cellcolor{gray!20}0.56\\ 
& \cellcolor{gray!20}1/8  & \cellcolor{gray!20}2.64 & \cellcolor{gray!20}292.92& \cellcolor{gray!20}0.82  &\cellcolor{gray!20}0.56 \\
\bottomrule
\end{tabular}
\vspace{-0.2cm}
\end{table}

\begin{table}[h]
\footnotesize
\caption{Quantitative comparison of KV cache compression methods for class-conditional and text-to-image generation tasks}
\label{baselinecompar}
\centering
\setlength{\tabcolsep}{5pt}
\begin{tabular}{ccccccc}
\toprule
\multirow{2}{*}{\textbf{Methods}} &
\multirow{2}{*}{\textbf{$\rho$}} &
\multicolumn{2}{c}{\textbf{LlamaGen-XL}} &
\multicolumn{2}{c}{\textbf{Janus-Pro-1B}} & \\
\cmidrule(lr){3-4} \cmidrule(lr){5-6}
 &&  \textbf{FID-50k} $\downarrow$ & \textbf{IS}$\uparrow$ & \textbf{FID-30k}$\downarrow$ & \textbf{CLIP} $\uparrow$  \\
\midrule
\textbf{Full}  & 1& 2.77 & 282.41 & 23.85 & 0.233 \\
\midrule
\multirow{2}{*}{\textbf{Random}}  &1/6 & 92.66 & 14.34& 99.73& 0.178\\
& 1/8& 123.53&8.42&117.42 &0.167\\
\midrule
\multirow{2}{*}{\textbf{Streaming}}  &1/6 & 3.09  &307.69 &24.80 &  0.236\\
& 1/8& 4.37 &297.08  &32.17  &  0.228\\
\midrule
\multirow{2}{*}{\textbf{H2O}}  & 1/6 & 3.06& 287.67 &25.87  & 0.225\\
& 1/8& 3.72 &279.05  &29.46  & 0.214 \\
\midrule
\multirow{2}{*}{\textbf{R-KV}}  & 1/6 & 2.77&276.56& 24.63&  0.232\\
& 1/8 & 3.45  & 271.29 & 25.55 &0.232   \\
\midrule

\multirow{2}{*}{\textbf{LineAR}}  & \cellcolor{gray!20}1/6& \cellcolor{gray!20}\textbf{2.68} & \cellcolor{gray!20}\textbf{279.17} &  \cellcolor{gray!20}\textbf{22.86}& \cellcolor{gray!20}\textbf{0.236} &\\

& \cellcolor{gray!20}1/8&  \cellcolor{gray!20}\textbf{2.95}  & \cellcolor{gray!20}\textbf{282.91}&\cellcolor{gray!20}\textbf{24.03} & \cellcolor{gray!20}\textbf{0.237}   & \\
\bottomrule
\end{tabular}
\end{table}

\textbf{Class-conditional Image Generation.}
We further evaluate LineAR on LlamaGen-XL and LlamaGen-XXL using the ImageNet dataset. As shown in Table\,\ref{llamagenresults}, LineAR consistently achieves comparable FID performance to the full-cache baseline, even when the cache budget ratio is reduced to 1/8 on both models. Notably, LineAR achieves lower FID scores (2.68 vs. 2.77) on LlamaGen-XL with a 1/6 budget ratio and shows improvements in both FID and Inception Score on LlamaGen-XXL with a 1/4 budget ratio. These results demonstrate that LineAR not only significantly improves the efficiency of autoregressive image generation but also reduces visual redundancy in cached tokens, thereby enhancing generation quality.

\begin{figure*}[htbp]
    \centering
    \begin{minipage}{0.25\textwidth}  
        \includegraphics[width=1\linewidth]{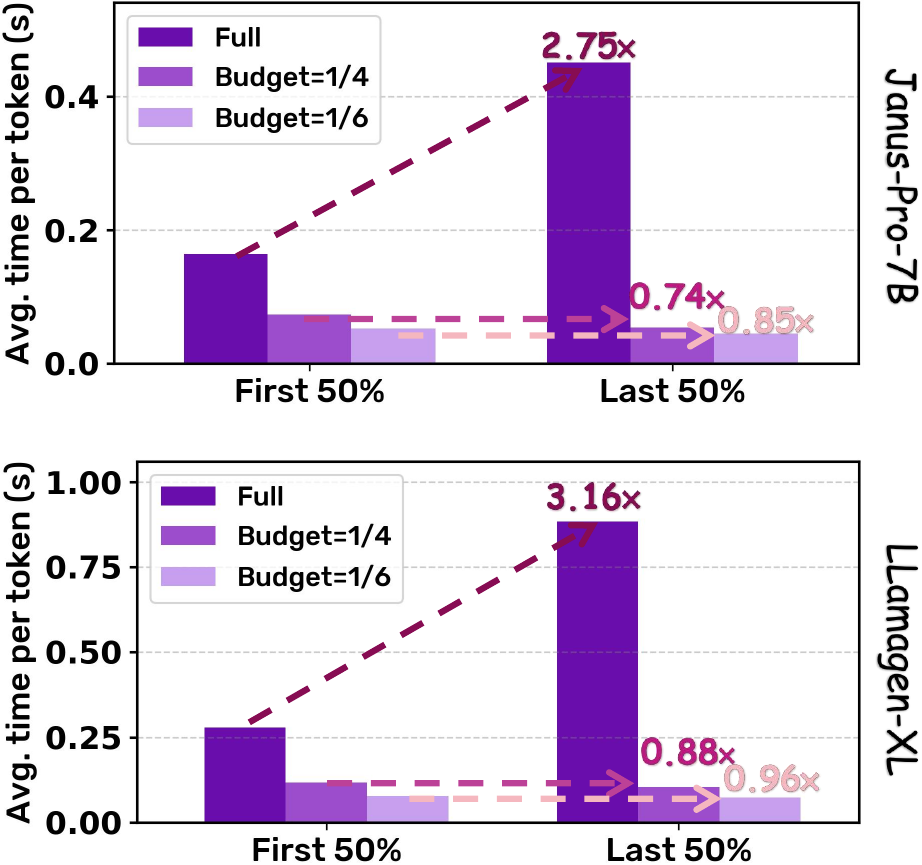} 
        \caption{Comparison of per-token generation speeds in the first and last 50\% of the Image.}
        \label{fig8}
    \end{minipage} \hfill
    \begin{minipage}{0.72\textwidth}  
    \vspace{-0.5cm}
        \begin{table}[H]
            \footnotesize
            \caption{Efficiency analysis on LlamaGen-XL, Janus-7B-Pro and Lumina-mGPT-768 models with lossless compression using LineAR, evaluated on NVIDIA A800 GPUs.}
            \label{efficiencyresults}
            \centering
            \setlength{\tabcolsep}{2.5pt}
    
\begin{tabular}{c cc cc cccc c}
\toprule 

\textbf{Models} & \textbf{Method} & \textbf{$\rho$}& \textbf{$B$} &\textbf{Batch} &\textbf{Memory} & \textbf{Mem. Saving}   & \textbf{Latency} & \textbf{Through.} &  \textbf{Speedup}\\
\midrule
\multirow{3}{*}{\thead{LlamaGen-XL \\($N=576$)}}&Full&1&576&256&64.75GB&-&331.52s&0.77$\text{it/s}$&1.00$\times$\\
\cmidrule(lr){2-10}
&\multirow{2}{*}{LineAR}&1/4&144&256&25.51GB&60.59\%&63.99s&4.00$\text{it/s}$&5.18$\times$\\
&&1/6&96&256&20.97GB&67.61\%&43.78s&5.85$\text{it/s}$&7.57$\times$\\
\midrule
\multirow{3}{*}{\thead{Janus-Pro-7B\\($N=576$)}}&Full&1&576&64&74.01GB&-&180.8s&0.35$\text{it/s}$&1.00$\times$\\
\cmidrule(lr){2-10}
&\multirow{2}{*}{LineAR}&1/4&144&64&47.59GB&35.70\%&40.3s&1.59$\text{it/s}$&4.49$\times$\\
&&1/6&96&64&44.66GB&39.66\%&32.19s&1.99$\text{it/s}$&5.62$\times$\\

\midrule
\multirow{3}{*}{\thead{Lumina-mGPT-768 \\($N=2352$)}}&Full&1&2352&12&55.48GB&-&283.99s&0.04$\text{it/s}$&1.00$\times$\\
\cmidrule(lr){2-10}
&\multirow{2}{*}{LineAR}&1/6&392&12&21.43GB&61.38\%&146.82s&0.08$\text{it/s}$&1.93$\times$\\
&&1/8&294&12&19.64GB&64.69\%&133.51s&0.09$\text{it/s}$&2.13$\times$\\

\bottomrule
\end{tabular}
        \end{table}
    \end{minipage}
\end{figure*}

\textbf{Comparison with Baselines.}
We compare LineAR with several representative KV cache compression methods originally developed for decoding-stage cache compression in LLMs, adapted here for AR image generation. These baselines include: StreamingLLM~\cite{xiao2023efficient}, which retains tokens at fixed intervals; H2O~\cite{zhang2024h2o}, which keeps persistently important tokens via accumulated attention; 
and R-KV~\cite{cai2025r}, which selects tokens based on contextual redundancy among reasoning contexts. 
A Random eviction baseline is also included for reference. 
Comparisons are performed under various cache budgets on both class-conditional (ImageNet) and text-to-image (MS-COCO) generation.
As shown in Table~\ref{baselinecompar}, LineAR consistently outperforms all baselines, achieving the best FID scores across tasks and budget ratios. 
Random eviction performs the worst, as indiscriminate removal discards critical initial anchors and recent-line tokens required for maintaining global style and local coherence. 
At a 1/6 budget ratio, all methods maintain relatively stable performance since they still preserve these key tokens to some extent, with R-KV ranking second due to its redundancy-aware design. 
However, under the more aggressive 1/8 ratio, all baselines suffer clear quality degradation, revealing their limited adaptation to visual-specific characteristics. 
In contrast, LineAR remains robust with minimal drop, demonstrating that its inter-line–guided eviction enables progressive, nearly lossless cache compression.

\subsection{Efficiency Analysis}
LineAR’s efficient management of the visual KV cache results in both memory savings and inference speedup across different architectures. Table\,\ref{efficiencyresults} presents the results under the lossless compression settings. Specifically, LineAR achieves up to 67.61\% memory savings on LlamaGen-XL with a 1/6 budget ratio, while significantly increasing throughput by up to $7.57\times$ compared to the full-cache baseline. On Janus-7B-Pro and Lumina-mGPT-768, LineAR also achieves 39.66\% and 61.38\% memory savings, respectively, along with $5.62\times$ and $2.13\times$ speedups at 1/6 and 1/8 budget ratios. Another advantage of LineAR is that it effectively mitigates the decline in visual autoregressive decoding throughput. Figure\,\ref{fig8} illustrates the average per-token generation speed comparison between the first and last 50\% of the decoding steps. Models with full-cache suffer from a throughput drop as the visual KV cache gradually accumulates, leading to much slower speed in the later generation. In contrast, LineAR maintains a stable decoding speed by limiting the KV cache within a fixed range, which is particularly beneficial for practical applications.

\begin{table}[!t]
\vspace{-0.2cm}
\centering
\footnotesize
\caption{Component alation on Janus-Pro-1B with $\rho=1/6$.}
\label{ablation_strategy}
\begin{tabular}{ccc|ccc}
\toprule
$\hat{\mathbf{P}}_{\text{init}}$ & $\hat{\mathbf{P}}_{\text{rec}}$ &
$\hat{\mathbf{P}}_{\text{mid}}$ & 
\textbf{FID-30k}$\downarrow$ & \textbf{CLIP}$\uparrow$ & \textbf{Geneval}$\uparrow$ \\

\midrule
& \ding{52}& \ding{52}& 203.38&0.156 &  0.04 \\
\ding{52}& &\ding{52} &79.96 & 0.191 & 0.13 \\
\ding{52}& \ding{52} & &24.80 & 0.236 &  0.67\\
\ding{52}&\ding{52} & \ding{52}&\textbf{22.86}  &\textbf{0.236} & \textbf{0.71}  \\
\bottomrule
\end{tabular}

\end{table}

\begin{table}[!t]
\centering
\footnotesize
\caption{Inter-line guidance albation on Janus-Pro-1B.}
\label{ablation_interline}
\begin{tabular}{cc|ccc}
\toprule
\textbf{LineAR} & \textbf{$\rho$} & 
\textbf{FID-30k}$\downarrow$ & \textbf{CLIP}$\uparrow$ & \textbf{Geneval}$\uparrow$ \\

\midrule
w. AttAcc &1/6&23.27&0.236&0.68\\
w. IL Guid. &1/6&\textbf{22.86} &\textbf{ 0.236}&\textbf{0.71}\\
w. AttAcc &1/8&24.30&0.237& 0.61\\
w. IL Guid. &1/8&\textbf{24.03} &\textbf{0.237}& \textbf{0.65}\\
\bottomrule
\end{tabular}
\vspace{-0.2cm}
\end{table}

\subsection{Ablation Study}
\textbf{Effectiveness of Compression Strategy Components.} 
We analyze the contributions of each component in our progressive compression strategy. As shown in Table\,\ref{ablation_strategy}, removing either the initial anchors $\hat{\mathbf{P}}_{\text{init}}$ or the recent lines' cache $\hat{\mathbf{P}}_{\text{rec}}$ significantly degrades image quality, emphasizing the need to preserve alignment with conditional tokens and local dependencies. Adding selective tokens $\hat{\mathbf{P}}_{\text{mid}}$ further improves performance, reducing FID-50k from 24.80 to 22.86, highlighting the importance of inter-line guided selection in retaining key tokens for subsequent line generation.

\textbf{Effectiveness of Inter-line Guided Eviction.}  
We evaluate the effectiveness of inter-line guided eviction (denoted as IL Guid.) by comparing it with the attention accumulation-based selection strategy (denoted as AttnAcc) under different budget ratios. Table\,\ref{ablation_interline} presents the results. Compared to the AttnAcc method, the model without inter-line guidance experiences significantly higher FID values under both budget ratios. This suggests that tokens evicted via inter-line guidance are less harmful to subsequent generations, leading to better progressive compression.

\begin{figure}[!t]
  \centering
    \vspace{-0.25cm}
  \includegraphics[width=1\linewidth]{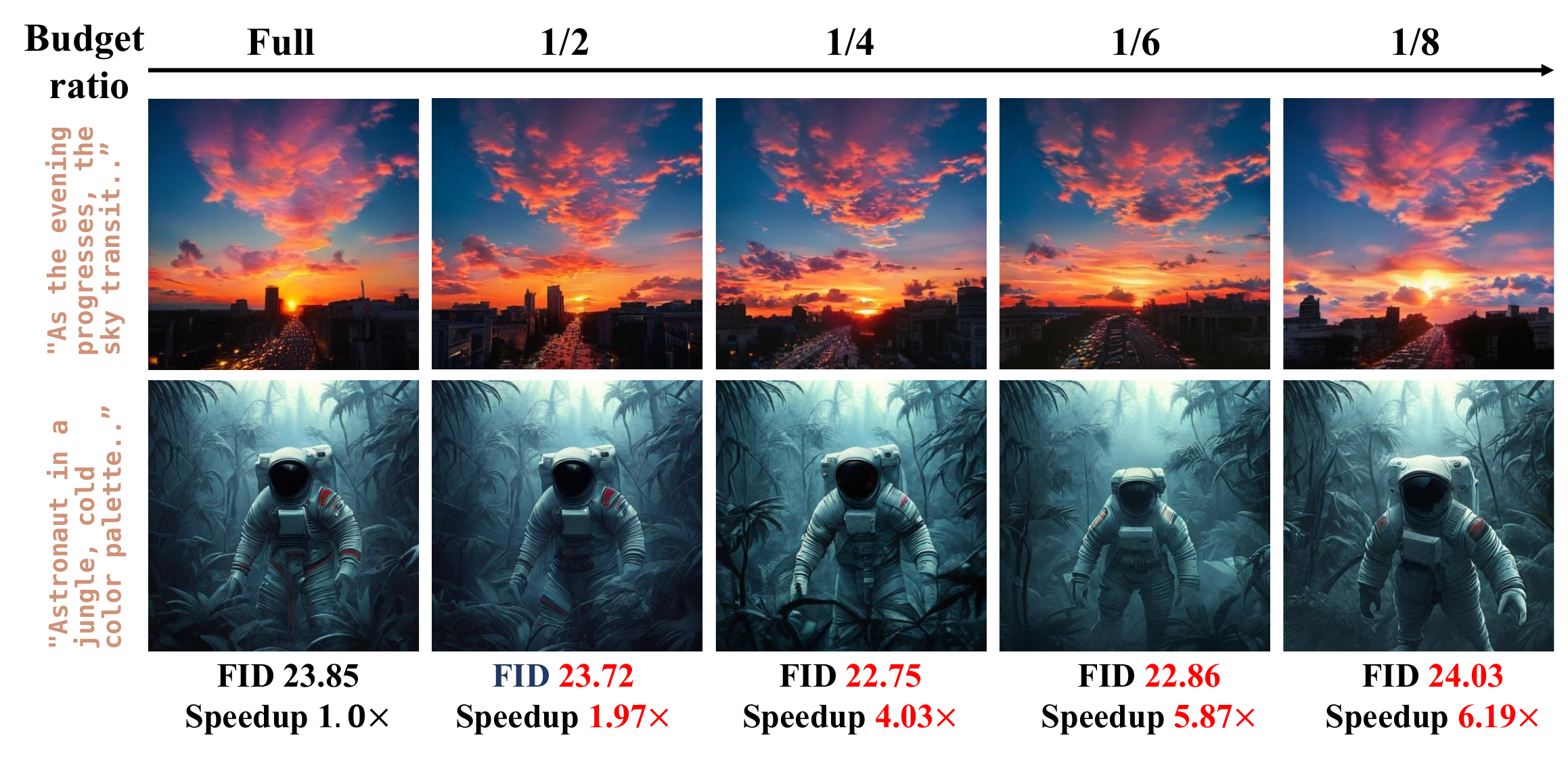}
    \caption{Comparison results under different budget ratios.}
  \label{ibr}
  \vspace{-0.3cm}
\end{figure}

\textbf{Impact of Budget Ratio.}
Figure\,\ref{ibr} presents both qualitative and quantitative results on Janus-Pro-1B under varying budget ratios. As the budget ratio decreases, there is a trade-off between FID scores and speedup. Moderate compression removes redundancy from visual attention, improving both quality and speed, while excessive compression results in slightly higher FID scores. Despite this, LineAR consistently delivers impressive performance, maintaining minimal quality loss and preserving the global style and color tones, closely aligned with the prompts.


\textbf{Additional Ablation Study.} Due to space constraints, ablations on hyperparameter sensitivity and more analysis are deferred to the \textit{supplementary material}

\section{Conclusion}
We present LineAR, a novel, training-free KV cache compression method designed for efficient autoregressive image generation. LineAR manages the visual KV cache at the line level and enables lossless progressive compression through inter-line guidance. It allows autoregressive image generation with only a few lines cached, achieving lossless performance. Extensive experiments across multiple AR image generation models demonstrate its effectiveness.
{
    \small
    \bibliographystyle{ieeenat_fullname}
    \bibliography{main}
}


\end{document}